\pdfoutput=1
\documentclass[11pt]{article}

\usepackage[final]{acl}
\usepackage{times}
\usepackage{latexsym}

\usepackage{graphicx}

\usepackage{makecell}
\usepackage{algorithm}
\usepackage{algorithmic}
\usepackage{amsmath}
\usepackage{cleveref}
\usepackage{array}
\usepackage{multirow}
\usepackage{booktabs}
\usepackage{amstext}
\newcommand{\mathcolorbox}[3]{\fcolorbox{#1}{#2}{$ #3$}}
\definecolor{box+}{cmyk}{0.00, 0.10, 0.20, 0.00}
\definecolor{line+}{cmyk}{0.00, 0.20, 0.93, 0.10}
\definecolor{box-}{cmyk}{0.00, 0.17, 0.18, 0.02}
\definecolor{line-}{cmyk}{0, 0.38, 0.39, 0.18}

\usepackage[T1]{fontenc}
\usepackage[utf8]{inputenc}

\usepackage{microtype}

\usepackage{inconsolata}
\title{\textit{Muffin}: Mitigating Unhelpfulness in Emotional Support Conversations\\with Multifaceted AI Feedback}

\author{Jiashuo Wang, Chunpu Xu, Chak Tou Leong, Wenjie Li\thanks{\ \ Corresponding author.}, Jing Li \\
  Hong Kong Polytechnic University\\
  \tt \{csjwang,cswjli,jing-amelia.li\}@comp.polyu.edu.hk\\
  \tt \{chun-pu.xu,chak-tou.leong\}@connect.polyu.hk
}

\begin{document}
\maketitle
\begin{abstract}
Emotional support conversation systems are designed to alleviate users' emotional distress and assist them in overcoming their challenges. 
While previous studies have made progress, their models occasionally generate unhelpful responses, which are intended to be supportive but instead have counterproductive effects.
Since unhelpful responses can hinder the effectiveness of emotional support, it is crucial to mitigate them within conversations.
Our solution is motivated by two principal considerations: (1) multiple facets of emotional support are expected to be considered when developing emotional support conversation models, and (2) directly reducing the probability of generating unhelpful responses can effectively mitigate their occurrence.
Accordingly, we introduce a novel \textbf{model-agnostic} framework named \underline{M}itigating \underline{u}nhelpfulness with multi\underline{f}aceted AI \underline{f}eedback for emot\underline{i}o\underline{n}al support (\textit{Muffin}).
It first employs a multifaceted AI feedback module designed to assess the helpfulness model responses across various facets of emotional support. 
Leveraging contrastive learning, Muffin then reduces the unhelpful responses' likelihoods.
To validate the effectiveness of our proposed framework, we apply Muffin to various previous emotional support generation models, including the state-of-the-art.
Experimental results demonstrate that Muffin can significantly mitigate unhelpful response generation while enhancing response fluency and relevance. We release our codes at \url{https://github.com/wangjs9/Muffin}.
\end{abstract}

\section{Introduction}\label{sec:introduction}
\begin{figure}[tb!]
\centering
\includegraphics[width=\linewidth]{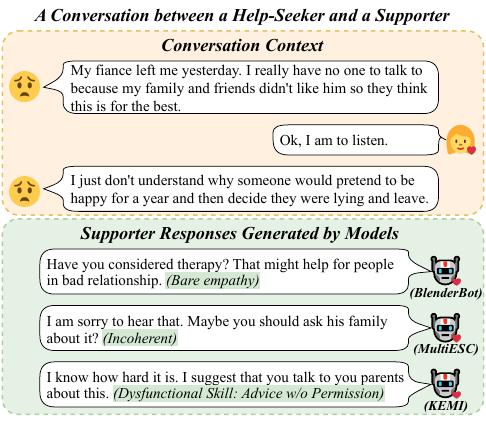} 
\caption{Examples of unhelpful responses generated by recent emotional support conversation models, including BlenderBot \cite{roller-etal-2021-recipes}, MultiESC \cite{cheng-etal-2022-improving}, and KEMI \cite{deng2023knowledge}.}
\label{fig:example}
\end{figure}

Emotional support conversation systems (supporters) are designed to generate responses that can buffer the emotional distress experienced by users (help-seekers) and help users to work through the challenges they are confronting \cite{liu-etal-2021-towards}. 
Recently, many studies have contributed to this field \cite{deng2023knowledge,zhao-etal-2023-transesc,cheng-etal-2022-improving,tu-etal-2022-misc}. 
Despite great success, their models occasionally generate well-intended responses that produce a counterproductive support effect, i.e., exacerbating the negative emotional states of users or inhibiting effective problem-solving, as shown in \Cref{fig:example}.
In the psychology and communication theories, these failed support attempts are termed ``\textit{unhelpful messages}'' \cite{greene2003handbook,burleson1985production}.

The frequency of unhelpful responses generated by some of the previous models is not extremely high,
e.g, approximately $30\%$ of responses generated by MultiESC \cite{cheng-etal-2022-improving} on ESConv benchmark \cite{liu-etal-2021-towards} are identified as unhelpful under strict evaluation criteria.
However, their occurrence can significantly undermine earlier supportive efforts and damage the trust between the help-seeker and the supporter \cite{llewelyn1988client}.
Therefore, mitigating models' generation of unhelpful responses is critical. 
We aim to address this problem with the following two deliberations.
\textit{\textbf{D1} - Consideration of Multiple Facets}: 
Many previous studies generate responses that primarily emphasize a single facet of emotional support, e.g., one of empathetic expression \cite{li2024enhancing}, communication skill efficiency \cite{cheng-etal-2022-improving,liu-etal-2021-towards}, or response coherence \cite{deng2023knowledge}, in each of their models.
However, such a singular emphasis on one facet often leads to the oversight of the others, potentially resulting in unhelpfulness \cite{greene2003handbook}, as exemplified in \Cref{fig:example}.
\textit{\textbf{D2} - Direct Minimization of Unhelpful Response Probability}:
Previous models are typically optimized by minimizing the negative log-likelihood of the golden responses. 
Moving beyond this optimization objective, we aim to specifically mitigate unhelpful responses by directly targeting and reducing the probability of their generation.

Accordingly, this paper introduces a novel \textbf{model-agnostic} framework called \underline{M}itigating \underline{u}nhelpfulness with multi\underline{f}aceted AI \underline{f}eedback for emot\underline{i}o\underline{n}al support (\textit{Muffin}).
For \textbf{\textit{D1}}, we design a multifaceted AI feedback module within the Muffin framework.
This module assesses whether a specific response is unhelpful from multiple facets of emotional support.
% , including empathetic expression, skill efficiency, and response coherence.These facets are commonly considered essential for supportive responses in previous studies.
Leveraging the advanced capabilities of recent large language models (LLMs), we implement this module by instruction-tuning LLaMA, avoiding inefficient and expensive human feedback collection.
Then, we continue optimizing an emotional support conversation model with its previous training objective and a new one to implement \textbf{\textit{D2}}.
The additional objective is to minimize the likelihood of unhelpful responses, which is implemented by contrasting unhelpful responses, identified by the feedback module, and the other (non-unhelpful) ones.
Through these two steps, we aim to mitigate the unhelpful responses generated by a given emotional support conversation model.
Experimental results highlight the effectiveness of our framework, demonstrating that Muffin can enhance the helpfulness of previous emotional support conversation models, including those recognized as state-of-the-art. The main contributions of this work are:
\begin{enumerate}
\item We recognize and address a crucial problem in recent emotional support conversation models, i.e., the generation of unhelpful responses, a key concern in effective emotional support.

\item We propose Muffin, a novel model-agnostic framework designed to mitigate unhelpful response generation. It incorporates a multifaceted AI feedback module to distinguish unhelpful generated responses and mitigates responses identified as unhelpful by leveraging contrastive learning.

\item We undertake experiments with the latest emotional support conversation models, including state-of-the-art ones, to demonstrate Muffin's effectiveness in mitigating the models' tendency to produce unhelpful responses.

\end{enumerate}
\section{Related Work}
\subsection{Emotional Support Conversation}
In the domain of emotional support conversation generation, prior studies have achieved some success.
Specifically, they have each emphasized and incorporated different facets of emotional support, such as empathetic expression, communication skills, and response coherence, into their respective models.
Some of them consider a single facet.
For example, MultiESC \cite{cheng-etal-2022-improving} only considers the communication skill efficacy of responses by planning response strategies.
KEMI \cite{deng2023knowledge} incorporates related external knowledge for response generation. Although the response coherence is enhanced, the efficacy of communication skills is ignored.
\citet{li2024enhancing} enhance the empathetic expression of LLMs through chain-of-though.
Beyond these, MISC \cite{tu-etal-2022-misc} generates supportive responses considering both commonsense and communication skills, thereby enhancing two facets of emotional support.
TranESC \cite{zhao-etal-2023-transesc} incorporates commonsense, communication skills, and emotional elements into response generation. 
Most all these models are optimized by minimizing the negative log-likelihood of golden responses, instead of the unhelpful response likelihood.
% Despite the advancement, most of the previous models occasionally generate unhelpful responses.
% Although the latter two methods consider more than one emotional support facet, neither minimizes the frequency of unhelpful responses directly.
% Our work proposes a novel approach that specifically tackles these two issues, aiming to enhance the generation of helpful responses.

\begin{figure*}[t]
\centering
\includegraphics[width=\textwidth]{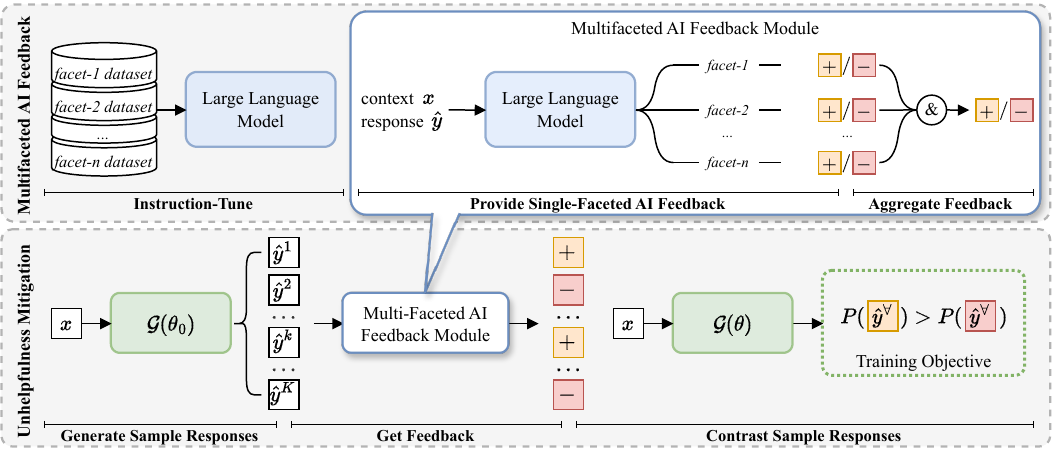} 
\caption{The overview of our proposed model-agnostic framework\textemdash Muffin. $\mathcolorbox{line+}{box+}{+}$ and $\mathcolorbox{line-}{box-}{-}$ indicate helpful (non-unhelpful) and unhelpful labels, respectively.}
\label{fig:model}
\end{figure*}

\subsection{Contrastive Learning for Text Generation}
Contrastive learning was initially employed to learn meaningful representations by contrasting positive and negative samples in the field of natural language processing \cite{logeswaran2018efficient,gutmann2012noise}. 
Recently, it has been applied to text generation \cite{jiashuo2023aligning,zheng-etal-2023-click,liu2022brio}, achieving impressive performance across various settings. 
During training, the model is exposed to a range of ``hard'' negative and positive samples through contrastive learning, enabling the model to distinguish preferred outputs from less desirable ones. 
Consequently, selecting positive and negative samples is crucial in this process. 
In this paper, responses generated by the model that are deemed unhelpful are negative samples, while all other responses are considered positive samples.
\section{Preliminaries}
\subsection{Unhelpfulness of Emotional Support}\label{sec:unhelpful}
We draw upon theories in psychology and communication \cite{greene2003handbook,burleson1985production} and adopt the term ``\textit{unhelpful}'' to characterize responses that consistently produce negative outcomes in emotional support conversations.
Conversely, responses that yield positive outcomes, or at the very least, do not cause negative effects, are termed ``\textit{helpful}.''
These theories also suggest that an unhelpful response often stems from a flaw in merely one specific facet of emotional support. 
Often, the flaw directly exacerbates the user's negative feelings or hinders effective problem-solving.
For instance, a response can be deemed unhelpful if it either neglects the individual's feelings and needs (lacking empathy) or portrays the individual's behavior as problematic (exhibiting a dysfunctional communication skill: confront). 
In our work, we use this feature to identify unhelpful responses.

\subsection{Problem Definition}
Our primary goal is to mitigate the generation of unhelpful responses.
Rather than training a new model from scratch, we aim to refine a pre-trained emotional support conversation model with the dataset it was originally trained on.
This process unfolds as follows.
Let $\mathcal{G}(\theta_0)$ represent the model trained on a dataset $\mathcal{D}$, where $\theta_0$ denotes the model parameters. 
Each instance in $\mathcal{D}$ is denoted as $(x, y)$, with $x$ as the input and $y$ as the expected output. 
Usually, $x$ is the conversation context, but it contains additional related information in some models.
Assume that there are $K$ samples $\{{\hat{y}^{1}, \cdots, \hat{y}^{K}}\}$ with labels $\{{\hat{l}^{1}, \cdots,\hat{l}^{K}}\}$.
These samples are the diverse beam search generation results of $\mathcal{G}(x;\theta_0)$.
As for the label $\hat{l}^{k}\in\{0,1\}$, it represents feedback to indicate whether the sample $\hat{y}^{k}$ is unhelpful ($\hat{l}^{k}=0$) or not unhelpful ($\hat{l}^{k}=1$).
Our objective is to refine the model's parameters $\theta$ such that the likelihood of generating unhelpful samples is reduced relative to helpful ones.
In this process, we only modify the training process, ensuring that the model's architecture and the inference mechanism remain untouched.
Moreover, our approach is model-agnostic. 
This implies that $\mathcal{G}(\theta_0)$ can be any deep learning model designed and trained for emotional support conversations.

\section{Method}
The overall framework of Muffin is outlined in \Cref{fig:model}.
It is composed of two principal components, each specifically designed for deliberations in \Cref{sec:introduction}: \textit{\textbf{D1}: Consideration of Multiple Facets} and \textit{\textbf{D2}: Direct Mimimization of Unhelpful Response Probality}, respectively.
The multifaceted AI feedback module aims to identify whether a response from multiple facets of emotional support is unhelpful.
The unhelpfulness mitigation module mitigates the likelihood of unhelpful responses by contrasting helpful and unhelpful responses.

\subsection{Multifaceted AI Feedback}
We distinguish whether a response is unhelpful from multiple facets.
However, collecting feedback from humans is inefficient and costly. 
In addition, recent large language models (LLMs), such as the GPT series \cite{ouyang2022training} and LLaMA \cite{touvron2023llama}, demonstrate remarkable natural language understanding capabilities. 
Therefore, we decide to obtain feedback from AI.

\paragraph{Instruction-tuning}
Prompt engineering provides a simple and straightforward approach to obtaining feedback from LLMs.
However, our experiments suggest that it is challenging to manifest the full potential of LLMs for emotional support without investing significant effort in prompt design, which will be detailed later.
As an alternative, we elicit the desired capabilities of the LLM via instruction tuning \cite{wei2021finetuned}.
Specifically, we design task descriptions and instructions tailored to classification tasks related to different emotional support facets.
We employ corresponding datasets for these tasks.
The prompt employed is illustrated in \Cref{fig:instruction}.
Notably, the response class can indicate whether the response is unhelpful regarding this facet.
During the training phase, all \textit{\underline{texts in italics enclosed within curly braces}} are provided. During inference, the model is expected to generate the {\textit{\underline{response class}}, based on the other \textit{\underline{italicized inputs within the curly braces}}.

\begin{figure}[ht]
\centering
\includegraphics[width=\linewidth]{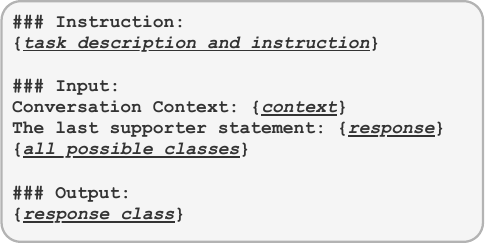} 
\caption{The prompt used by the Multifaceted AI Feedback for classifying the supporter's response.}
\label{fig:instruction}
\end{figure}

\paragraph{Multifaceted AI feedback module}
The final feedback for an emotional support response is derived from an aggregation of AI feedback across multiple facets.
For the given response and its context, we use the instruction-tuned LLM to provide feedback on all these facets respectively.
If feedback from any of these facets suggests the response is unhelpful, the response is accordingly labeled as $0$; otherwise, the response is deemed helpful (non-unhelpful) and labeled as $1$, as mentioned in \Cref{sec:unhelpful}.

\subsection{Unhelpfulness Mitigation}\label{sec:mitigation}
We mitigate $\mathcal{G}(\theta_0)$ generating unhelpful responses by contrasting helpful and unhelpful responses generated by $\mathcal{G}(\theta_0)$ itself, which can be implemented by the following three steps:

\paragraph{Generating sample responses}
We utilize $\mathcal{G}(\theta_0)$ to generate responses on its own training dataset $\mathcal{D}$ using diverse beam search \cite{vijayakumar2016diverse}.
Thus, for each instance $(x,y)\in\mathcal{D}$, there are $K$ sample responses $\{\hat{y}^{1},\cdots,\hat{y}^{K}\}$.

\paragraph{Getting feedback}
These responses can be generated because they have relatively high generation probabilities.
However, some of them can be unhelpful responses.
Therefore, we adopt the multifaceted AI feedback module to identify whether these responses are unhelpful. Thus, we obtain $K$ labels $\{\hat{l}^1\cdots,\hat{l}^{K}\}$, where $\hat{l}^k\in\{0,1\}$.

\paragraph{Contrasting sample responses}
We expect that the model $\mathcal{G}$ can sign a higher likelihood to the helpful responses than the unhelpful ones.
Therefore, we contrast them using the following loss:
\begin{equation}\label{eqn:clloss}
\begin{aligned}
\mathcal{L}_{cl}=&\frac{1}{2K}\sum_{i}\sum_{j\neq i}\max(0,\\
&-(\hat{l}^i-\hat{l}^j)\times(\mathrm{P}(\hat{y}^{i}|x)-\mathrm{P}(\hat{y}^{j}|x)+\lambda)),
\end{aligned}
\end{equation}
where $\lambda$ is the margin hyperparameter. 
Moreover, $\mathrm{P}(\hat{y}^{i}|x)$ is the length-normalized log-probability of the response $\hat{y}^{i}$, and it is be computed by:
\begin{equation}
    \mathrm{P}(\hat{y}^{i}|x)=\sum_{t=1}^{|\hat{y}^i|}\frac{\log\mathcal{G}(\hat{y}^i_t|x,\hat{y}^i_{<t};\theta)}{|\hat{y}^i|^\alpha},
\end{equation}
where $\alpha$ is the length penalty hyperparameter.
In addition to the above loss, we also consider the negative log-likelihood loss to prevent the model's generation from deviating too much from the ground truth. 
The loss can be formulated as:
\begin{equation}
    \mathcal{L}_{gen}= - \frac{1}{|y|} \sum_{t=1}^{|y|}\log{\mathcal{G}(y_t|x,y_{<t};\theta)}.
\end{equation}
The final loss is the combination of the above two losses:
\begin{equation}\label{eqn:totalloss}
    \mathcal{L} = \beta_{cl}\mathcal{L}_{cl} + \beta_{gen}\mathcal{L}_{gen},
\end{equation}
where $\beta_{cl}$ and $\beta_{gen}$ are weight hyperparameters. 

\section{Experiments}\label{sec:experiments}
% In this section, we first introduce the experimental setups for the multifaceted AI feedback module, the base model $\mathcal{G}(\theta_0)$, and emotional support conversation dataset $\mathcal{D}$. Then, we show the performance of components in Muffin, including the multifaceted AI feedback module and the response generation, respectively. Then, detailed analyses are displayed.
\subsection{Experimental Setups}
\paragraph{Base models ($\mathcal{G}(\theta_0)$)}
Our proposed method, i.e., Muffin, is a model-agnostic approach designed to mitigate the unhelpfulness of an existing emotional support conversation model. 
To examine its effectiveness, we experiment with five recent models:
BlenderBot (Vanilla) \cite{roller-etal-2021-recipes}, BlenderBot-Joint (Joint) \cite{liu-etal-2021-towards}, MultiESC \cite{cheng-etal-2022-improving}, TransESC \cite{zhao-etal-2023-transesc}, and KEMI \cite{deng2023knowledge}.
We obtain each model's parameters $\theta$ using its official implementation and the default hyperparameters.

When training Muffin$_{\mathcal{G}(\theta_0)}$, we try to use the same training hyperparameters as the base model $\mathcal{G}(\theta_0)$, including batch size and random seed.
However, we use a smaller learning rate, i.e., $3\times10^{-5}$, to help the model converge more efficiently. 
We set the epoch number as $1$ since the training loss converged within one epoch.
The margin hyperparamter $\lambda$ is $0.01$, the length penalty hyperparameter $\alpha$ is $1$, and the weight hyperparamters $\beta_{cl}$ and $\beta_{gen}$ are both $1$.
In the response generation phase, as described in \Cref{sec:mitigation}, we set the number of sample responses $K$ to $10$. 
Consequently, both the beam size and the number of beam groups are configured to be $10$, while all other generation hyperparameters are the same as its base model.

\paragraph{ESConv dataset ($\mathcal{D}$)}
The ESConv dataset \cite{liu-etal-2021-towards} is used to train the aforementioned base models. 
ESConv is a benchmark for emotional support conversations, comprising approximately 1K conversations with 31K utterances. 
All base models, with the exception of TransESC, follow the original division of ESConv for training, validation, and testing, using an 8:1:1 ratio. 
TransESC employs a random split while maintaining the same ratio. 
Notably, each model adopts different data preprocessing methods. 
We adhere to each base model's specific data division and pre-processing.

\subsection{Facets of Emotional Support}\label{sec:experiment_facets}
We consider three essential facets of emotional support: empathetic expression, skill efficiency, and response coherence, which the base models incorporate into their models.
Here, we would like to describe the unhelpfulness of each facet and detail the corresponding classification dataset.

\paragraph{Empathetic expression}
Empathetic expressions signify the supporter's interest in and comprehension of the help-seeker's perspective. 
Conversely, their absence can impede conversation engagement and obstruct establishing a trust-based relationship between the supporter and the help-seeker \cite{morse1992beyond}. 
While empathy encompasses various aspects \cite{wang2021empathetic,paiva2017empathy,bohart1997empathy}, we adopt the comprehensive framework proposed by \cite{sharma2020computational}.
It identifies three empathy communication mechanisms, i.e., emotional reaction, interpretations, and explorations, each assessed across three levels: no communication, weak communication, and strong communication.
Our work considers responses exhibiting no empathy across all mechanisms unhelpful.
For example, ``sleeplessness can result in upsetness,'' which inappropriately offers mere information, is considered unhelpful in empathetic expression, especially when responding to a statement like ``I am upset''.

For training and testing LLMs in classifying unhelpful responses regarding empathetic expression, we also utilize the dataset compiled by \citet{sharma2020computational}. 
It consists of $3$K context-response pairs.
Each response within this dataset is assessed based on three previously mentioned empathy communication mechanisms.
Responses consistently labeled as ``no communication'' across all these mechanisms are identified as unhelpful.

\paragraph{Skill efficiency}
The applications of effective strategies can help supporters convey appropriate and impactful support messages \cite{greene2003handbook}, deepening the understanding of the help-seeker's state and facilitating problem solution \cite{hill2009helping}.
However, some dysfunctional skills lead to opposite effects \cite{barsky2001evaluating,burleson1985consistencies}.
For example, while \textit{advice} is typically seen as beneficial, \textit{advice without permission} can be less effective than employing no specific skills at all. 
This study assesses skill efficiency using motivational interviewing skill codes \cite{moyers2003motivational}, which include three general categories: MI adherent, MI non-adherent, and others. 
Responses classified as MI non-adherent category are deemed unhelpful.

We utilize the Motivational Interviewing (MI) dataset proposed by \citet{welivita-pu-2022-curating} to classify unhelpful responses in the context of skill efficiency. 
This dataset comprises $17$K context-response pairs, where each response is annotated with one of three MI codes: MI adherent, MI non-adherent, or others. 
Responses labeled as ``MI non-adherent'' are considered unhelpful.

\paragraph{Response coherence}
While response coherence is a fundamental expectation in almost all conversational systems \cite{huang2020challenges}, it holds particular significance in emotional support conversation systems.
Incoherent responses can confuse and impede effective communication.
We categorize responses as coherent or incoherent, with the latter labeled unhelpful.

We have synthesized a dataset, derived from the base model's training set $\mathcal{D}$, specifically for the purpose of response coherence detection.
Specifically, we randomly selected $4$K context-response pairs from $\mathcal{D}$. 
For each pair, the original response is categorized as coherent. 
To introduce incoherence, we employ two methods: firstly, by selecting a response from a different conversation, and secondly, by modifying keywords or important information in the original response to create a subtly incoherent variant.
These methods result in a total of approximately $12$K context-response pairs, which are then utilized for classifying unhelpful responses in terms of their response coherence.

\subsection{Multifaceted AI Feedback}
\paragraph{Instruction-tuning settings}
We instruction-tune a 7B LLaMA \cite{touvron2023llama} for the multifaceted AI feedback module and use a low-rank adaptor (LoRA) \cite{hu2021lora} for efficiency.
Specifically, we freeze LLaMA's weight and inject trainable rank decomposition matrices into query, key, value, and output layers.
The learning rate is $3\times10^{-4}$, and the training epoch is $12$.
We merge the datasets described in \Cref{sec:experiment_facets}, formatting each context-response pair according to the instruction template presented in \Cref{fig:instruction}. 
Consequently, this merged dataset encompasses a total of $22$K instances, which are utilized for instruction-tuning. 
The dataset is partitioned into training, validation, and test sets following an 8:1:1 split ratio.

\begin{figure}[t]
\centering
\includegraphics[width=\linewidth]{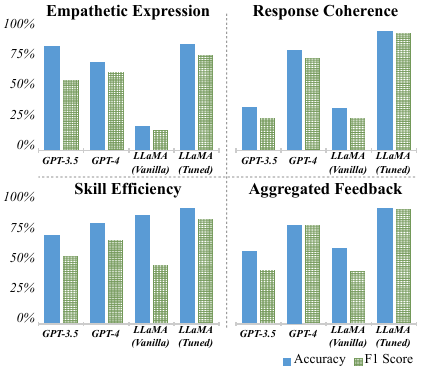}
\caption{Comparison of performance among various Language Model Models (LLMs) including GPT-3.5, GPT-4, LLaMA (Vanilla), and LLaMA (Tuned) in classification tasks related to different facets of emotional support, as well as the aggregated feedback.}
\label{fig:feedback}
\end{figure}

\paragraph{Module performance}
The instruction-tuned LLaMA model is subsequently employed within the Multifaceted AI Feedback Module. 
The module's effectiveness is evaluated using the instruction-tuning test set, with the results depicted in Figure \ref{fig:feedback}. 
We report both accuracy and F1 scores for data instances throughout the entire testing set and across distinct facets. 
Additionally, we extend our evaluation to include the module's performance when utilizing three other LLMs: GPT-3.5, GPT-4, and the original (vanilla) LLaMA model.
Without fine-tuning, GPT-4 outperforms the other two models in terms of the classification effectiveness. 
Upon closer examination, GPT-3.5 is prone to classifying responses as non-unhelpful. 
The vanilla LLaMA model frequently classifies responses into a singular category across different facets, leading to the worst performance. 
However, after instruction-tuning, LLaMA exhibits a significant enhancement in performance, with an accuracy of $90.72\%$ and an F1 score of $89.86\%$ on the aggregated feedback, thereby exceeding the capabilities of its counterparts. 
This remarkable improvement provides strong justification for our decision to apply the 7B instruction-tuned LLaMA within the Multifaceted AI Feedback Module.

\begin{table*}[t]
    \centering
    \small
    \renewcommand\arraystretch{0.95}
    \begin{tabular}{c|cccc|c|c|c|c}
    \hline
    \textbf{Model}
    & \textbf{B-1} & \textbf{B-2} & \textbf{B-3} & \textbf{B-4} & \textbf{R-L} & \textbf{METEOR} & \textbf{CIDEr} & \textbf{Extreme} \\
    \hline    
    Vanilla & 18.23$^{~}$ & 7.02$^{~}$ & 3.49$^{~}$ & 1.99$^{~}$ & 16.09$^{~}$ & 7.31$^{~}$ & 14.95$^{~}$ & 50.48$^{~}$ \\
    Muffin$_{\textbf{Vanilla}}$ & 19.43$^{*}$ & 7.58$^{*}$ & 3.66$^{~}$ & 2.02$^{~}$ & 16.26$^{~}$ & 7.72$^{*}$ & 13.90$^{*}$ & 51.00$^{*}$\\
    \hline
    Joint & 18.77$^{~}$ & 7.54$^{~}$ & 3.79$^{~}$ & 2.15$^{~}$ & 17.72$^{~}$ & 7.59$^{~}$ & 17.38$^{~}$ & 50.96$^{~}$\\
    Muffin$_{\textbf{Joint}}$ & 20.59$^{*}$ & 8.38$^{*}$ & 4.26$^{*}$ & 2.54$^{*}$ & 18.35$^{*}$ & 8.18$^{*}$ & 19.12$^{*}$ & 51.46$^{*}$ \\
    \hline
    TransESC & 17.32$^{~}$ & 7.10$^{~}$ & 3.63$^{~}$ & 2.18$^{~}$ & 17.47$^{~}$ & 7.53$^{~}$ & 22.07$^{~}$ & 51.33$^{~}$ \\
    Muffin$_{\textbf{TransESC}}$ & 17.19$^{~}$ & 7.17$^{*}$ & 3.73$^{*}$ & 2.25$^{*}$ & 17.54$^{*}$ & 7.58$^{*}$ & 22.72$^{*}$ & 51.57$^{*}$  \\
    \hline
    KEMI & 19.85$^{~}$ & 8.15$^{~}$ & 4.24$^{~}$ & 2.52$^{~}$ & 17.17$^{~}$ & 7.92$^{~}$ & 15.09$^{~}$ & 50.85$^{~}$ \\
    Muffin$_{\textbf{KEMI}}$ & 20.01$^{*}$ & 8.31$^{*}$ & 4.36$^{*}$ & 2.60$^{*}$ & 17.30$^{*}$ & 7.99$^{*}$ & 15.45$^{*}$ & 51.11$^{*}$   \\
    \hline
    MultiESC & 21.79$^{~}$ & 9.19$^{~}$ & 4.98$^{~}$ & 3.05$^{~}$ & 20.92$^{~}$ & 8.93$^{~}$ & 28.84$^{~}$ & 52.59$^{~}$ \\
    Muffin$_{\textbf{MultiESC}}$ & 21.83$^{*}$ & 9.28$^{*}$ & 5.12$^{*}$ & 3.21$^{*}$ & 21.26$^{*}$ & 8.92$^{~}$ & 31.26$^{*}$ & 52.83$^{~}$ \\
    \hline
    \end{tabular}
    \caption{Automatic evaluation results and AI feedback from the multifaceted AI feedback module. For all metric scores and feedback, a higher value indicates better performance. The values marked with $*$ indicate the results are statistically significant with $p<0.05$.}
    \label{tab:auto}
\end{table*}
\subsection{Emotional Support Response Generation}
\paragraph{Automatic evaluation}
We evaluate the quality of emotional support responses by a range of automatic evaluation metrics, including BLEU \cite{papineni2002bleu} (\textit{B-1/2/3/4}), ROUGE (\textit{R-L}) \cite{lin2004rouge}, \textit{METEOR} \cite{banerjee2005meteor}, \textit{CIDEr} \cite{vedantam2015cider}, and BOW Embedding-based matching score (\textit{Extreme}) \cite{liu2016not}. 
These metrics are good at evaluating the similarity between the generated response and the ground truth.

\Cref{tab:auto} showcases the performance of Muffin with different base models in all automatic evaluation metrics. 
In general, Muffin demonstrates significant enhancements across nearly all evaluation metrics.
Moreover, it can be observed that the performance of Muffin$_{\mathcal{G}(\theta_0)}$ is predominantly influenced by its base model $\mathcal{G}(\theta_0)$, assessed through automatic evaluations. 

% However, there are two interesting phenomena.
% The first is that BlenderBot-Vanilla achieves a relatively high score in terms of communication skill feedback, despite being the sole model without a dedicated mechanism to integrate communication skills. 
% Upon closer examination, we discern that this model frequently generates responses such as ``I can understand that...'' or ``I've experienced something similar...''. 
% Approximately $17\%$ of all its generated responses contain such content, in contrast to the $9.34\%$ prevalence in the ESConv dataset and roughly $10\%$ in BlenderBot-Joint responses.
% \footnote{Such responses are counted via regular expression matching.}
% These responses are typified as self-disclosure. 
% While this communication skill does not undermine prior emotional support efforts, making it undetectable as ``unhelpful'' by the AI feedback module, a conversation dominated by self-disclosure responses does not fully meet the objective of providing support.
% Another strange thing is that TransESC obtains extremely low AI feedback in Empathy, although it designs an emotion transition mechanism.
% However, according to findings in TransESC's ablation experiment \cite{zhao-etal-2023-transesc}, the annotated emotion labels in training data and model-generated emotional knowledge contain noise.
% This could be the possible reason for the low Empathy feedback score.

\paragraph{Human evaluation}
Following previous work \cite{deng2023knowledge,liu-etal-2021-towards}, we compare the base model and its corresponding Muffin model on five aspects, which are
(1) \textit{Fluency}: which model's response is more fluent?
(2) \textit{Identification}: which model's response is more skillful in identifying the user's problem?
(3) \textit{Comforting}: which model's response is better at comforting the user?
(4) \textit{Suggestion}: which model can give more helpful and informative suggestions?
(5) \textit{Helpfulness}: which model's response is generally more helpful from the aspect of the help-seeker?
Specifically, for each $\mathcal{G}(\theta_0)$-Muffin$_{\mathcal{G}(\theta_0)}$ pair, we randomly select 100 instances for comparison.
Then, we ask four unique human evaluators to vote which response is better.
They can select ``tie'' if responses are considered equal.
We average their results as the final result.

\Cref{fig:human} summarizes the A/B test results on BlenderBot-Joint \cite{liu-etal-2021-towards}, KEMI \cite{deng2023knowledge}, and MultiESC \cite{cheng-etal-2022-improving}, along with their corresponding Muffin models. 
% \footnote{
These three settings are selected for their significant performance in automatic evaluation.
% }. 
The inter-rater agreement, i.e., Fleiss' Kappa \cite{fleiss1971measuring}, is $0.39$, implying fair agreement.
Our Muffin models are regarded as more helpful in general, as evidenced by their higher \textit{Helpfulness}. 
Responses generated by Muffin models are slightly more fluent than those generated by base models. 
We also observe that 'ties' are common in evaluations of response fluency, mainly because the responses generated are typically fluent. 
Compared with the corresponding base model $\mathcal{G}(\theta_0)$, the Muffin$_{\mathcal{G}(\theta_0)}$ model shows some more powerful capability in identifying the help-seeker's problem.
Moreover, Muffin$_{\mathcal{G}(\theta_0)}$ models can generate responses that have better effects to comfort the users than $\mathcal{G}(\theta_0)$. 
Annotators also prefer responses generated by Muffin$_{\mathcal{G}(\theta_0)}$ because of their more helpful and informative suggestions.
These results prove that Muffin indeed mitigates the unhelpfulness of emotional support conversation models.

\begin{figure}[t]
\centering
\includegraphics[width=\linewidth]{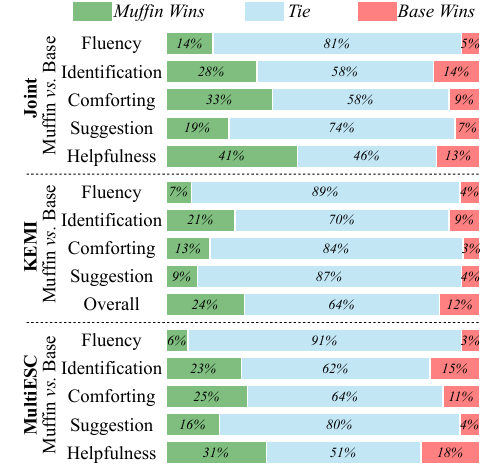}
\caption{Human A/B test results. Displayed within each bar, from left to right, are the ratios for ``\textit{Muffin Wins}'', ``\textit{Tie}'', and ``\textit{Base Wins}''.}
\label{fig:human}
\end{figure}

\begin{table*}[htp]
\centering
 \begin{minipage}[t]{0.45\textwidth}
    \centering
    \small
    \renewcommand\arraystretch{1}
    \begin{tabular}{c|cccc}
    \hline
    \multicolumn{5}{c}{\textit{A. Generating One Response}}\\
    \hline
    \textbf{Model} & \textbf{\textit{emp.}} & \textbf{\textit{skill}} & \textbf{\textit{cohr.}} & \textbf{\textit{agg.}} \\
    \hline    
    Vanilla & 81.22 & 90.43 & 80.69 & 64.83 \\
    Muffin$_{\textbf{Vanilla}}$ & 85.83 & 92.82 & 84.33 & 71.26 \\
    \hline
    Joint & 80.61 & 88.93 & 80.65 & 61.48 \\
    Muffin$_{\textbf{Joint}}$ & 82.33 & 90.47 & 83.04 & 64.76 \\
    \hline
    TransESC & 81.06 & 91.44 & 74.76 & 63.24 \\
    Muffin$_{\textbf{TransESC}}$ & 81.28 & 91.49 & 78.67 & 66.02 \\
    \hline
    KEMI & 83.01 & 88.47 & 85.76 & 70.90 \\
    Muffin$_{\textbf{KEMI}}$ & 83.40 & 88.68 & 87.15 & 72.33 \\
    \hline
    MultiESC & 83.39 & 90.66 & 85.38 & 70.00 \\
    Muffin$_{\textbf{MultiESC}}$ & 85.24 & 92.92 & 86.06 & 72.38 \\
    \hline
    \end{tabular}
    \label{tab:appn_facetA}
  \end{minipage}
  \begin{minipage}[t]{0.45\textwidth}
    \centering
    \small
    \renewcommand\arraystretch{1}
   \begin{tabular}{c|cccc}
    \hline
    \multicolumn{5}{c}{\textit{B. Generating Ten Responses}}\\
    \hline
    \textbf{Model} & \textbf{\textit{emp.}} & \textbf{\textit{skill}} & \textbf{\textit{cohr.}} & \textbf{\textit{agg.}} \\
    \hline    
    Vanilla & 81.90 & 92.03 & 78.89 & 64.51 \\
    Muffin$_{\textbf{Vanilla}}$ &  89.84 & 96.05 & 82.60 & 73.78 \\
    \hline
    Joint & 79.31 & 90.56 & 79.75 & 63.30 \\
    Muffin$_{\textbf{Joint}}$ & 83.43 & 92.26 & 80.02 & 66.54\\
    \hline    
    TransESC & 74.16 & 92.78 & 43.93 & 41.21 \\
    Muffin$_{\textbf{TransESC}}$ & 74.20 & 92.92 & 44.15 & 41.43\\
    \hline
    KEMI & 81.16 & 87.49 & 81.57 & 66.23 \\
    Muffin$_{\textbf{KEMI}}$ & 81.39 & 87.64 & 81.61 & 66.47  \\
    \hline
    MultiESC & 78.61 & 89.67 & 77.33 & 60.83 \\
    Muffin$_{\textbf{MultiESC}}$ & 78.77 & 89.73 & 77.26 & 60.87 \\
    \hline
    \end{tabular}
   \end{minipage}
   \caption{The AI feedback is sourced from the comprehensive AI feedback module. The left subtable showcases the percentage of each model's helpful (non-unhelpful) responses employing the decoding strategy of the base model; the right subtable displays the percentage of each model's helpful responses utilizing diverse beam search with a beam size set at $10$. \textit{\textbf{emp.}}, \textit{\textbf{skill}}, \textit{\textbf{cohr.}} and \textit{\textbf{agg.}} represent empathetic expression, skill efficiency, response coherence, and aggregated feedback, respectively. All values are expressed in percentages (\%), where higher percentages signify superior performance. The values in the left subtable are statistically significant with $p<0.05$.}
   \label{tab:facet}
\end{table*}
\begin{table*}[t]
    \centering
    \small
    \begin{tabular}{c|cccc|c|c|c|c}
    \hline
    \textbf{Model}
    & \textbf{B-1} & \textbf{B-2} & \textbf{B-3} & \textbf{B-4} & \textbf{R-L} & \textbf{METEOR} & \textbf{CIDEr} & \textbf{Extreme} \\
    \hline
    Joint & 18.77 & 7.54 & 3.79 & 2.15 & 17.72 & 7.59 & 17.38 & 50.96 \\
    Muffin$_\textbf{Joint}$ & 20.59 & 8.38 & 4.26 & 2.54 & 18.35 & 8.18 & 19.12 & 51.46  \\
    \hline
    Muffin$_\textbf{Joint}$ (\textit{emp.}) & 19.58 & 8.06 & 4.04 & 2.33 & 18.46 & 7.87 & 19.61 & 51.43 \\
    Muffin$_\textbf{Joint}$ (\textit{skill}) & 18.68 & 7.51 & 3.81 & 2.19 & 17.98 & 7.67 & 17.84 & 50.98 \\
    Muffin$_\textbf{Joint}$ (\textit{cohr.}) & 20.04 & 8.10 & 4.04 & 2.26 & 18.24 & 7.95& 18.07 & 51.31\\
    \hline
    \end{tabular}
    \caption{Ablation study results. Muffin$_\textbf{Joint}$ ($X$) indicates the mitigation process only uses AI feedback in terms of the facet $X$. All values are statistically significant with $p<0.05$.}
    \label{tab:ablation}
\end{table*}

\paragraph{Helpfulness Evaluation via AI Feedback}
We provide the multifaceted AI feedback results as a reference, as outlined in \Cref{tab:facet}.
We utilize the multifaceted AI feedback module to identify various models' helpful (non-unhelpful) responses and compute their percentage, displayed in the left subtable.
Furthermore, we analyze the helpful response percentage when each model generates ten responses using diverse beam search, reported in the right subtable.\footnote{The assertion ``approximately 30\% of responses generated by MultiESC on the ESConv benchmark are identified as unhelpful'' is derived from the findings presented in this table.}

Overall, Muffin demonstrates enhancements in AI feedback across multiple facets. However, three intriguing phenomena emerge.
\textbf{(1).} While the left subtable indicates an increase in the frequency of helpful responses attributable to the Muffin framework, the evidence presented in the right subtable is weaker (\textit{\textbf{agg.}}), particularly when the base model is TransESC, KEMI, or MultiESC. 
This finding aligns with our loss function (\Cref{eqn:clloss} and \Cref{eqn:totalloss}). 
We have introduced a contrastive loss to the original generation loss.
This loss does not significantly mitigate unhelpfulness. Instead, it assigns higher generation probabilities to helpful (non-unhelpful) responses. 
Consequently, when generating one response, the output is the one with the highest probability. 
\textbf{(2).} Despite lacking a dedicated mechanism for incorporating communication skills, BlenderBot-Vanilla attains a notably high score in communication skill efficiency (\textbf{\textit{skill}}). 
Upon closer examination, we observe that this model frequently produces responses such as ``I can understand that...'' or ``I've experienced something similar...''.
These responses are categorized as self-disclosure. 
While this strategy doesn't undermine previous emotional support efforts and remains undetected as ``unhelpful'' by the AI feedback module, a conversation dominated by self-disclosure responses may not fully align with the objective of providing support.
\textbf{(3).} Another noteworthy observation is that TransESC receives exceptionally low AI feedback in response coherence (\textbf{\textit{cohr.}}), despite its incorporation of external knowledge. 
However, insights from TransESC's ablation experiment \cite{zhao-etal-2023-transesc} suggest that the generated knowledge may contain noise. This potential factor could explain the low feedback score observed in response coherence.

\begin{table}[t]
    \footnotesize
    \centering
    \renewcommand\arraystretch{1}
    \begin{tabular}{m{.23\hsize}<{\centering}m{.67\hsize}}
    \hline
    \multicolumn{2}{m{.95\hsize}<\centering}{\textit{\textbf{Conversation History}}} \\ 
    \hline
    \multicolumn{2}{m{.95\hsize}}{
    \colorbox{pink!55}{Help-seeker}: I'm an alcoholic. I get that thought in my head and its like I become obsessed.} \\ 
    
    \multicolumn{2}{m{.95\hsize}}{
    \colorbox{pink!55}{Supporter}: Why not try to put attention on some other things instead of a beer?}\\
    
    \multicolumn{2}{m{.95\hsize}}{
    \colorbox{pink!55}{Help-seeker}: Hmmm like what?} \\
    \hline
    \multicolumn{2}{m{.95\hsize}<\centering}{\textit{\textbf{Model Generated Responses}}} \\ 
    \hline
    Joint & You can try to be normal. That is a great idea. \colorbox{cyan!27}{(\textit{Lacking empathy})} \\
    Muffin$_{\text{Joint}}$ & Yes, there are some things that you can do, but you have to be aware of your own mental health, that's why it's important to know you are doing the right thing.\\
    \hline
    MultiESC & It sounds like you have a lot on your plate and you want to be normal. \colorbox{cyan!27}{(\textit{Inconsistent})} I can understand how that would make you feel.\\
    Muffin$_{\text{MultiESC}}$ & I think you should focus on yourself instead of a beer. Do you have any hobbies that you enjoy?\\
    \hline
    KEMI & I think you should try to find some new things to do to relax.\\
    Muffin$_{\text{TransESC}}$ & Have you tried meditation? It can help you relax and relax.\\ 
    \hline
    \end{tabular}
    \caption{Comparison of responses generated by BlenderBot-Joint, MultiESC, and KEMI with their respective Muffin-enhanced counterparts.}
    \label{tab:case}
\end{table}

\paragraph{Ablation study}
To assess the impact of different facets of feedback, we undertake an ablation study. 
Specifically, we employed AI feedback from each individual facet to distinguish helpful from unhelpful responses, subsequently contrasting them to mitigate unhelpfulness.
This study is conducted using BlenderBot-Joint as the base model, given its relatively straightforward architecture and the notable improvements exhibited by Muffin$_\text{Joint}$.

The findings, as presented in \Cref{tab:ablation}, reveal that relying solely on AI feedback from a single facet for unhelpfulness mitigation results in diminished performance in automatic evaluations when compared to the comprehensive Muffin$_\text{Joint}$ model. 
This underscores that considering multiple facets of emotional support when building emotional support conversation models (\textbf{\textit{D1}}) is necessary. 

Another insight from \Cref{tab:ablation} is that all three ablated models outperform the base model.
It suggests that the proposed solution, directly mitigating unhelpful responses (\textbf{\textit{D2}}), is reasonable and effective.
Moreover, the results also indicate that the quality of helpful and unhelpful responses will influence the effects of unhelpfulness mitigation.
Another finding is that the impact of different facets on overall performance varies, a trend we consider to be expected and rational. Specifically, by combining the ablation study results presented in \Cref{tab:ablation} with the findings in \Cref{tab:facet}, we observe different percentages of non-unhelpful responses across different facets. Notably, the facet of skill efficiency exhibits the lowest percentage of non-unhelpful responses, leading to the relatively poorer performance of Muffin (skill) in the ablation study. Conversely, the percentages of non-unhelpful responses for other facets are more comparable, resulting in similar performance levels for Muffin (emp.) and Muffin (cohr.). This analysis underscores the nuanced influence of different facets on the efficacy of our approach.
 
\paragraph{Case study}
To intuitively illustrate the superiority of Muffin over its base model, \Cref{tab:case} presents a comparative case study, comparing responses generated by three prominent base models and their corresponding Muffin versions.
From the comparison of BlenderBot-Joint and Muffin$_\text{Joint}$, we can observe that the BlenderBot-Joint implies that the help-seeker can be abnormal now.
Such a statement ignores the help-seeker's feelings, barely expressing empathy.
For the comparison of responses generated by BlenderBot-Joint and Muffin$_\text{Joint}$, the former tends to state facts more directly, subtly implying that the help-seeker might be experiencing an abnormal state.
Such a statement ignores the help-seeker's feelings, barely expressing empathy.
In contrast, Muffin$_\text{Joint}$ conveys concern for the help-seeker's well-being and attempts to solve the problem by shifting the help-seeker's perspective, amplifying the empathetic undertone.
In the case of MultiESC, Muffin$_\text{MultiESC}$ crafts a response that aligns more closely with the context, addressing the inconsistency of the response generated by MultiESC.
Lastly, comparing KEMI with Muffin$_\text{KEMI}$, even though KEMI's response does not exhibit glaring issues, the Muffin version stands out as more beneficial. 
This distinction arises because Muffin$_\text{KEMI}$, in contrast to KEMI's general advice, offers a more specific and actionable recommendation, aligning closely with the help-seeker's request for precise advice.
\section{Conclusion}
In this work, we focus on mitigating unhelpful responses generated by recent emotional support conversation models.
Such unhelpful responses, despite their well-intentioned nature, can inadvertently counteract prior supportive efforts. 
Analyzing the potential causes for unhelpful responses, we introduce a novel model-agnostic framework Muffin. 
In specific, it contains a multifaceted AI feedback module, which can discern helpful and unhelpful responses generated by a specific emotional support conversation model.
Then Muffin contrasts helpful and unhelpful responses generated by this model, in order to reduce the likelihood of the unhelpful responses.
Experimental results underscore Muffin's efficacy, showcasing enhancements in both automatic evaluations and human ratings. 
This suggests that Muffin can mitigate helpfulness in emotional support conversations.

\newpage
\section*{Limitations}
Although Muffin's effectiveness is apparent, opportunities for further refinement exist. 
Our present efforts address general cases of unhelpfulness, specifically targeting responses generally perceived as unhelpful by most help-seekers. 
Nonetheless, it is crucial to acknowledge that certain responses may adversely affect particular individuals under specific circumstances. 
Consequently, this underscores the need for personalizing the emotional support conversation system to meet individual user requirements.
However, this personalization presents substantial challenges. Collecting and processing personal data raises serious privacy and security concerns. 
Striking a balance between effective personalization and the potential for privacy intrusion remains a delicate issue. 
Furthermore, reconciling the goals of personalizing individual user experiences with the need to generalize models across a broader user base poses a fundamental conflict in system design.

\section*{Ethical Considerations}
In our experiments, we adopted open-sourced datasets, including ESConv \cite{liu-etal-2021-towards}, empathetic classification dataset \cite{sharma2020computational}, and MI dataset \cite{welivita-pu-2022-curating}.
All personally identifiable information was removed from these datasets.
For the human ratings, we emphasized the comfort and well-being of our annotators.
Moreover, our research explores the development of emotional support conversation systems. 
Compared with existing methods, our proposed approach represents a significant leap towards establishing a more secure emotional support conversation framework. 
Consequently, we confidently assert that our research is conducted in strict adherence to the ethical guidelines prescribed by the Association for Computational Linguistics (ACL).

\section*{Acknowledgements}
This work was supported by the Research Grants Council of Hong Kong (PolyU/5204018, PolyU/15207920, PolyU/15213323) and National Natural Science Foundation of China (62076212).
% Bibliography entries for the entire Anthology, followed by custom entries
%\bibliography{anthology,custom}
% Custom bibliography entries only
\bibliography{custom}

% \newpage

\appendix
\section{Multifacetd AI Feedback Module}
This section provides more details about the multifaceted AI feedback module.
\begin{table*}[tb]
    \small
    \centering
    \renewcommand\arraystretch{1.25}
    \begin{tabular}{m{.2\hsize}<{\centering}|m{.35\hsize}|m{.35\hsize}}
    \hline
    \textbf{Facet} & \makecell[c]{\textbf{Context}} & \makecell[c]{\textbf{Unhelpful Response}}\\
    \hline
    \multirow{5}{*}{\textbf{\textit{Empathetic Expression}}} & \textit{Help-seeker}: Just wanna sleep. Can't motivate at all. I don't know what to do. & \textit{Supporter}: As Nas says, sleep is the cousin of death. It's the kindest form of oblivion.\\ 
    \cline{2-3}
    & \textit{Help-seeker}: What do you want to be when you grow up? I don't know what to do with my life or what I truly love when all I really want is not to exist. & \textit{Supporter}: But what does fully grown up mean? I have my own house and a job etc. Am I fully grown up now? \\
    \hline
    \multirow{6}{*}{\makecell{\textbf{\textit{Communication Skill}}\\\textbf{\textit{Efficiency}}}} & \textit{Help-Seeker}: I do nothing but think about how I'm worthless and how I shouldn't be here. I've never tried or contemplated suicide. I've always wanted to fix my issues, but I never get around to it. How can I change my feeling of being worthless to everyone? & \textit{Supporter}: Still, you should watch out for this because it can be a sign of worsening depression.\\
    \cline{2-3}
    & \textit{Help-Seeker}: I try to have a ``heart-to-heart'' with my mom, but it usually winds up being messy and frustrating afterward. Thank you for your advice and words of encouragement. Sorry if that sounds silly. & \textit{Supporter}: Stop being so afraid...you really have nothing to lose :) Well, you are right.\\
    \hline
    \multirow{8}{*}{\textbf{\textit{Response Coherence}}} & \textit{Help-seeker}: Is it afternoon where you are? I'm still waking up; I should have more coffee. And I am just always tired of everything these days. & \multirow{6}{*}{\makecell[l]{\textit{Supporter}: Could you please explain why\\you are enjoying yourself?}}\\
    & \textit{Supporter}: Yes, what happened? Why are you tired? Please can you explain? Everything going ok?& \\
    & \textit{Help-seeker}: I think it's my depression. I have a very hard time getting motivated to do anything. Nothing seems worth it. & \\
    \cline{2-3}
     & \textit{Help-seeker}: I am upset. & \textit{Supporter}: Everyone can be happy. Can you tell me more?\\ 
    \hline
    \end{tabular}
    \caption{Unhelpful responses in terms of each facet of emotional support.}
    \label{tab:unhelpful_facets}
\end{table*}

\subsection{Multiple Facets of Emotional Support}\label{appn:facets}
For further clarification and better understanding, we have provided additional examples of unhelpful response instances related to the emotional support facets we emphasize in our work. 
These examples can be found in \Cref{tab:unhelpful_facets}.
Furthermore, we have expanded on the explanations of helpfulness and unhelpfulness in relation to each facet and the corresponding classification dataset.

\paragraph{Empathetic Expression}
\begin{figure*}[tb]
\centering
\includegraphics[width=\textwidth]{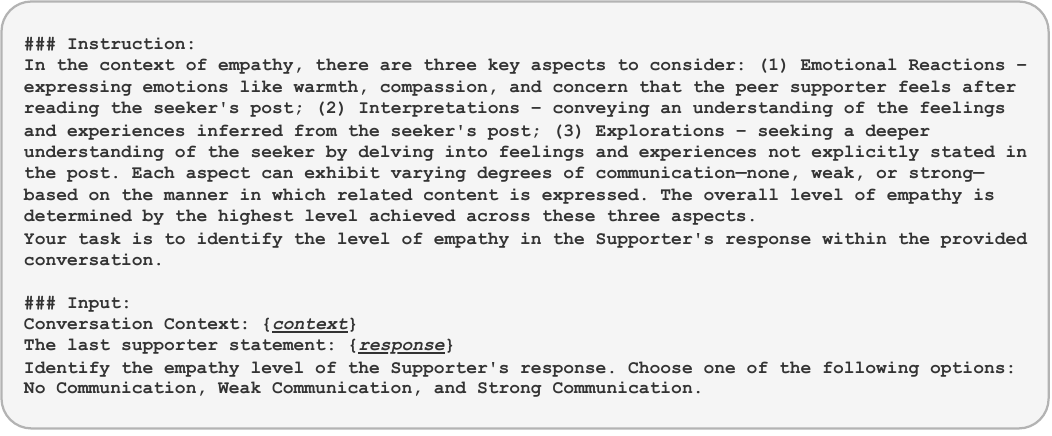} 
\caption{The specific prompt utilized by the Multifaceted AI Feedback for classifying supporter statements considering empathetic expression.}
\label{fig:empathetic_instruct}
\end{figure*}
The empathetic expression framework introduced by \citet{sharma2020computational} encompasses three empathy communication mechanisms, each illustrating a different approach to conveying empathy:
\begin{itemize}
\item \textbf{Emotional Reaction:} This mechanism involves expressing warmth, compassion, and concern for the help-seeker. An example is the statement ``I am here to help you,'' which demonstrates empathy through emotional reaction.
\item \textbf{Interpretations:} This approach reflects an understanding of feelings and experiences as inferred from the help-seeker's statements. For instance, ``I understand that you feel sad'' exemplifies the supporter's interpretations.
\item \textbf{Explorations:} Responses in this category explore feelings and experiences not explicitly stated by the help-seeker. An example is the close question ``Why are you feeling alone right now?''
\end{itemize}
In the framework and associated dataset outlined by \citet{sharma2020computational}, responses that solely offer advice (e.g., ``ask your school counselor what resources they can provide''), merely present factual information (e.g., ``mindful meditation helps overcome anxiety''), or are offensive or abusive (e.g., ``I don't know what to do; I am also feeling suicidal right now'') are considered non-empathetic and characterized as ``no communication.'' 
However, as noted by \citet{greene2003handbook}, information and suggestions can often be helpful, particularly when the help-seeker requires. 
Consequently, in our work, responses containing pertinent information or suggestions are classified as ``weak/strong communication'' if they align with the help-seeker's request in the context. 
To identify such responses, we utilize GPT-4 to detect contexts where the help-seeker explicitly (e.g., ``any suggestions'') or implicitly (e.g., ``I don't know what to do'') requests information or suggestions. 
Subsequently, these responses are manually evaluated to determine if they provide reasonable information or suggestions. 
For the classification task, the specific prompt is depicted in \Cref{fig:empathetic_instruct}.

\paragraph{Communication Skill Efficiency}
\begin{figure*}[tb]
\centering
\includegraphics[width=\textwidth]{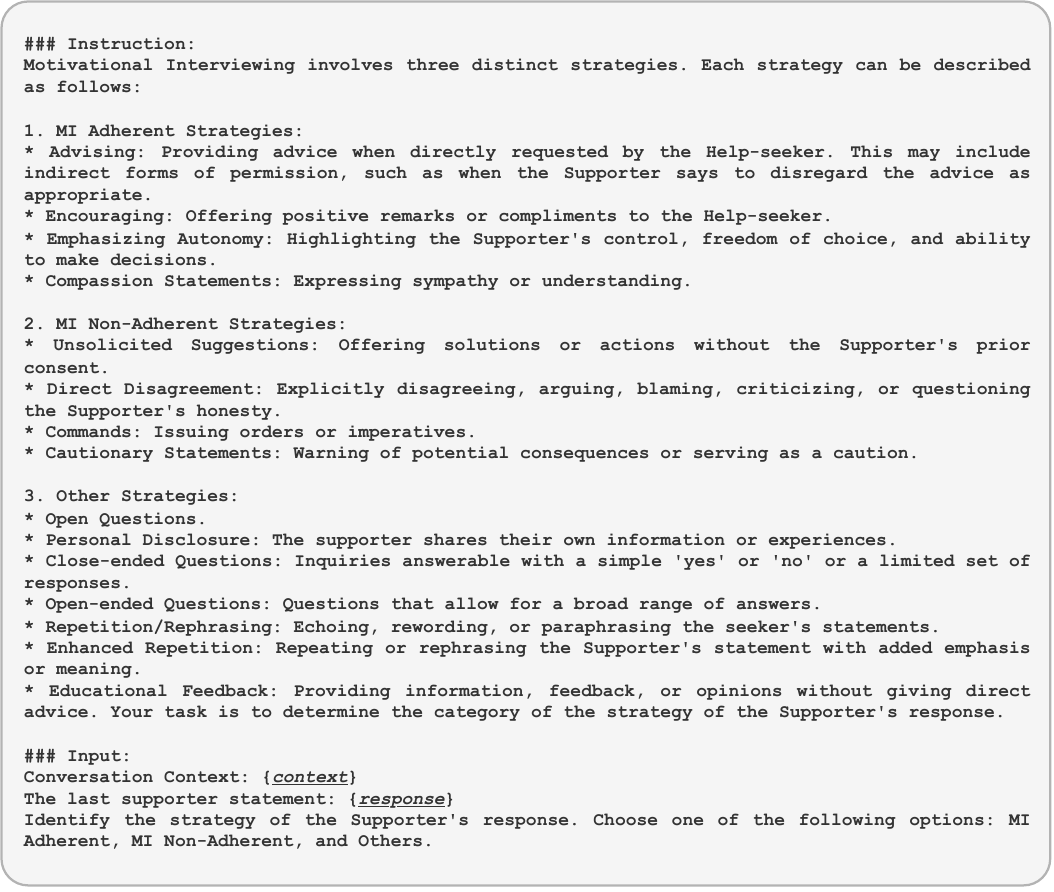} 
\caption{The specific prompt used by the Multifaceted AI Feed-
back for classifying the supporter’s statement regarding communication skill efficiency.}
\label{fig:skill_instruct}
\end{figure*}
We adopt the  Motivational Interviewing Treatment Integrity (MITI) code \citet{moyers2003motivational}, a well-established behavioral coding system that differentiates between favorable and unfavorable responses, to categorize responses into three classes based on their communication skills:
\begin{itemize}
\item \textbf{MI Adherent:} Responses in this category support help-seekers with empathetic and compassionate statements, fostering a sense of being heard, respected, and understood. For example, the response ``Well, there is really a lot going on for you right now'' qualifies as MI Adherent.
\item \textbf{Others:} This class encompasses responses that might not immediately elicit a positive support effect. 
These include responses like a closed question, such as ``Did you eat five fruits and vegetables this week?'' 
In our work, we regard both these types of responses and those classified as MI Adherent as \textit{helpful}. 
Our rationale is that such responses, though seemingly less impactful initially, can be beneficial in the long term. 
They contribute to the progression of the conversation and strengthen the relationship between the help-seeker and the supporter.
\item \textbf{MI Non-Adherent:} This category includes responses that involve arguing, confronting, or offering unsolicited advice, which may lead to resistance and impede problem-solving for help-seekers. An example is the response ``Yes, you are an alcoholic. You might not think so, but you are,'' which is classified as MI Non-Adherent.
\end{itemize}
To classify supporter statements based on communication efficiency, we utilize the MI dataset as proposed by \citet{welivita-pu-2022-curating}. Additionally, \Cref{fig:skill_instruct} illustrates the specific prompt employed for this task.

\paragraph{Response Coherence}\label{appn:cohe}
To develop the dataset for response classification in the aspect of response coherence, we initially selected approximately $4$K context-response pairs from the ESConv training dataset as the foundation for constructing the coherence data. 
It is important to note that this training dataset represents the intersection of the training sets used by all base models.
As mentioned in \Cref{sec:experiment_facets}, two types of incoherent variants were created: one by replacing the original response with a response from a different dialogue, and the other by altering keywords in the original response. 
For the latter approach, we utilized GPT-4 to modify keywords or crucial information, ensuring that the altered response either conveyed a contrary meaning or addressed a different topic.
\Cref{tab:coherence_instance} presents an illustrative example of this process.
The specific prompt employed for this task is illustrated in \Cref{fig:coherence_instruct}.
\begin{table}[tbh]
    \centering
    \small
    \renewcommand\arraystretch{1.25}
    \begin{tabular}{m{.19\hsize}<{\centering}|m{.69\hsize}}
    \hline
    \multicolumn{2}{c}{Context} \\ 
    \hline
    \multicolumn{2}{m{.94\hsize}}{\textit{Help-seeker}: Is it afternoon where you are? I'm still waking up; I should have more coffee. And, I am just always tired, tired of everything these days.} \\
    \hline
    Original Response  & Yes, what happened? Why are you tired? Please can you explain?\\
    \hline
    \multirow{4}{*}{\makecell{Incoherent\\Responses}} & I am glad to hear you are feeling a little better! Yes, bad management is so toxic. Even with great coworkers in a job you love, horrible management can ruin it quickly. (From another dialogue)\\
    \cline{2-2}
    & Why are you asking me to explain when you are feeling exhausted? Can you share your thoughts on this matter? (Keywords changed)\\
    \hline
    \end{tabular}
    \caption{The instance of a coherent response and its two incoherent variants.}
    \label{tab:coherence_instance}
\end{table}

\begin{table*}[tbh]
    \centering
    \small
    \renewcommand\arraystretch{1.25}
    \begin{tabular}{m{.2\hsize}<{\centering}|m{.11\hsize}<{\centering}|m{.1\hsize}<{\centering}m{.1\hsize}<{\centering}m{.15\hsize}<{\centering}m{.15\hsize}<{\centering}}
    \hline
    \multicolumn{2}{c|}{\textbf{Model}} & GPT-3.5 & GPT-4 & LLaMA (Vanilla) & LLaMA (Tuned) \\
    \hline
    \multirow{2}{*}{\textbf{\textit{Empathetic Expression}}} & \textit{accuracy} & 81.88 & 69.58 & 19.09 & 83.50 \\
    & \textit{F1 Score} & 54.89 & 61.58 & 16.03 & 74.12\\
    \hline
    \multirow{2}{*}{\textbf{\textit{Skill Efficiency}}} & \textit{accuracy} & 69.56 & 79.01 & 84.96 & 90.43\\
    & \textit{F1 Score} & 53.22 & 66.14 & 45.93 & 81.83\\
    \hline
    \multirow{2}{*}{\textbf{\textit{Response Coherence}}} & \textit{accuracy} & 33.47 & 78.76 & 33.31 & 92.98\\
    & \textit{F1 Score} & 25.26 & 72.62 & 24.98 & 92.06\\
    \hline
    \multirow{2}{*}{\textbf{\textit{Aggregated Feedback}}} & \textit{accuracy} & 57.11 & 78.01 & 59.08 & 90.72\\
    & \textit{F1 Score} & 41.99 & 77.19 & 40.96 & 89.86\\
    \hline
    \end{tabular}
    \caption{Performance of various Language Model Models (LLMs) on unhelpful response classification in terms of different facets and the aggregated decision. All values are expressed as percentages (\%).}
    \label{tab:feedback}
\end{table*}

\begin{figure*}[tb]
\centering
\includegraphics[width=\textwidth]{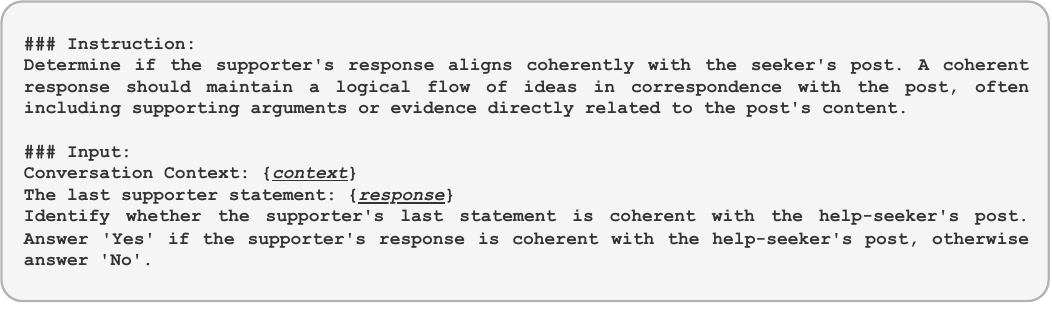} 
\caption{The specific prompt utilized by the Multifaceted AI Feedback for classifying the supporter's statement in terms of response coherence.}
\label{fig:coherence_instruct}
\end{figure*}

\subsection{Model Tuning and Module Performance}
We instruction-tuned LLaMA\footnote{\url{https://huggingface.co/decapoda-research/llama-7b-hf}} to equip the model with the capability required for unhelpful response classification tasks.\footnote{The implementation referred to is available at \url{https://huggingface.co/decapoda-research/llama-7b-hf}} Furthermore, we initialized the LoRA weights using a low-rank adapter that was fine-tuned on the Stanford Alpaca dataset.\footnote{\url{https://huggingface.co/tloen/alpaca-lora-7b}} 
We evaluated the capabilities of GPT-3.5 and GPT-4 by invoking the gpt-3.5-turbo and gpt-4 models, respectively, through their API\footnote{\url{https://platform.openai.com/docs/api-reference/chat/create}} during the period from December 2023 to January 2024. 
We set the temperature parameter to $0$ to ensure deterministic output generation. 
The detailed results of this evaluation are detailed in Table \ref{tab:feedback}.

GPT-3.5 refers to considering the supporter's response not unhelpful.
Prior to instruction-tuning, LLaMA exhibited notably poor performance across all facets of the classification tasks. The model consistently favored specific classes, such as ``Strong Empathy,'' ``Other,'' and ``Yes,'' corresponding to empathetic expression, communication efficiency in skill, and response coherence, respectively.

We also conduct human evaluation to examine the module. In particular, we randomly sample $200$ responses from the ESConv dataset and model-generated responses. 
For each response, an annotator was asked to assess whether it is unhelpful.
We compared the annotations with the multifaceted AI feedback, and the consistency rate was $88\%$.
We found that the multifaceted AI module is stricter than human annotators.
It is evident by the fact that for instances in which the multifaceted AI feedback and human annotations are different, the multifaceted AI feedback tends to consider the response unhelpful.

\begin{figure*}[tb]
\centering
\includegraphics[width=\textwidth]{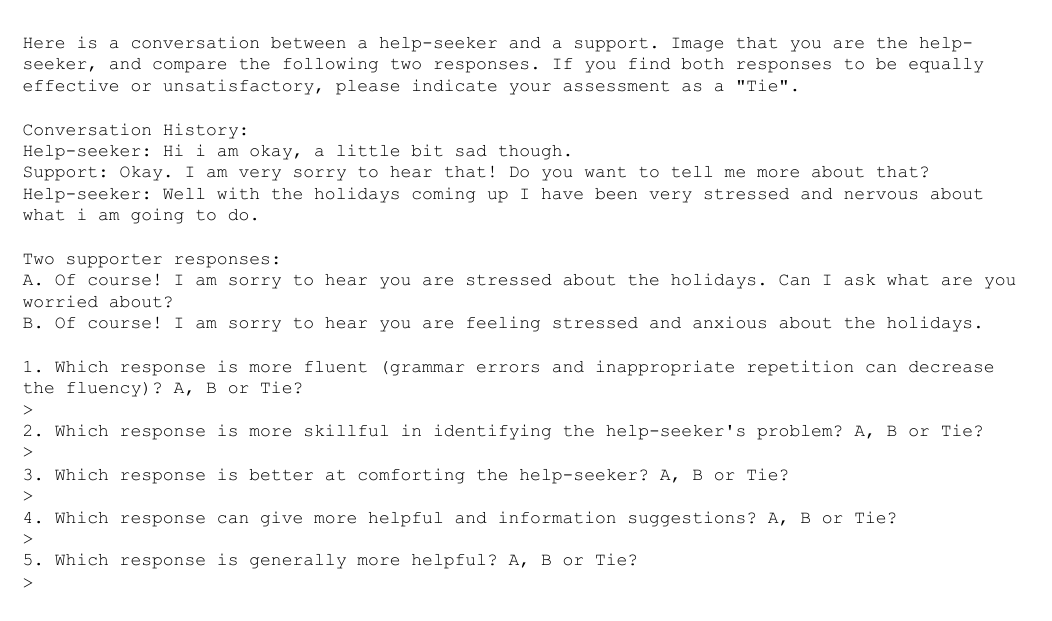} 
\caption{An example of an instance presented to annotators for evaluation.}
\label{fig:human_rating}
\end{figure*}

\section{Experimental Setup}
\subsection{Devices and Environment}
We employed Pytorch\footnote{\url{https://pytorch.org/}} for the implementation of all models, and conducted our experiments using Nvidia GeForce RTX 3090 GPUs.

\subsection{ESConv Dataset}
The ESConv dataset comprises a collection of emotional support conversations, each facilitated between a help-seeker and a supporter. 
While previous studies have partitioned and processed this dataset in various ways, our implementations of Muffin strictly follow each of their methodologies to maintain fairness.
However, it is important to note that direct comparisons of performance across these models may not be entirely fair due to the differences in their data preprocessing approaches.

\subsection{Base Models}
\paragraph{Vanilla} is a vanilla BlenderBot \cite{roller-etal-2021-recipes} trained on the dataset ESConv. We used the small version\footnote{\url{https://huggingface.co/facebook/blenderbot\_small-90M}} of BlenderBot in experiments following previous studies.

\paragraph{Joint} is built upon the backbone of BlenderBot and is specially trained to generate responses along with an expected communication skill at the beginning of each response \cite{liu-etal-2021-towards}.\footnote{\url{https://github.com/thu-coai/Emotional-Support-Conversation}} The focused facet of this model is skill efficiency. 

\paragraph{MultiESC} \cite{cheng-etal-2022-improving}\footnote{\url{https://github.com/lwgkzl/multiesc}} is an emotional support conversation model that mainly focuses on the communication skill efficiency. It predicts the strategies (communication skills) in the next several turns considering the the user's state.

\paragraph{KEMI} incorporates various knowledge for a mixed-initiative conversation model, which can provide emotional support \cite{deng2023knowledge}.\footnote{\url{https://github.com/dengyang17/KEMI}} 
Consequently, the response coherence is also enhanced.

\paragraph{TransESC} \cite{zhao-etal-2023-transesc}\footnote{\url{https://github.com/circle-hit/TransESC}} predicts the transitions of the user's emotion, the communication skills, and the conversation keywords. Then, such information is used for response generation. This model takes into account more than one facet of emotional support; however, it is not optimized to reduce the likelihood of unhelpful responses like other models.

\subsection{Human Evaluation}
For each pair of $\mathcal{G}(\theta_0)$ and Muffin$_{\mathcal{G}(\theta_0)}$, we randomly selected $100$ response pairs generated by both $\mathcal{G}(\theta_0)$ and Muffin$_{\mathcal{G}(\theta_0)}$ under identical conversational contexts. 
Selecting $100$ responses aligns with the standard practice for human evaluation in dialogue tasks, such as the experiments of TransESC \cite{zhao-etal-2023-transesc} and KEMI \cite{deng2023knowledge}. 
To prevent annotators from identifying the generation model based on the order of sentences, the sequence in which these two responses are presented is randomized for each evaluation.

Initially, the annotators were briefed about the nature of emotional support conversations to ensure a comprehensive understanding of the task's objectives. 
During the rating process, they were suggested to imagine themselves as the help-seekers within the conversations. 
After being provided with the conversation context, the annotators then proceeded to compare two generated responses, shown as \Cref{fig:human_rating}. 
Moreover, we prioritized the comfort and well-being of our annotators, advising them to pause or cease the annotation process if they encountered any content that made them feel uncomfortable.
Annotators were paid at a rate of 1.5$\sim$2 times their local hourly minimum wage. 
Based on annotators' feedback, it was estimated that approximately 40 seconds were spent on evaluating each response pair.

The inter-rater agreement among annotators, as measured by Fleiss's Kappa, is $0.39$. This value is relatively high, especially when compared to inter-rater reliability values in most subjective tasks, which typically fall within the range of 0.2 to 0.6 \cite{wong2021cross, cowen2017self}.
In the process of aggregating the annotations, we determine the winning response based on the consensus of annotators. 
If two annotators prefer response A, and another two annotators prefer a tie, we categorize response A as the winner. 
When an equal number of annotators favor two responses, we label the result as a tie.

\section{Additional Experimental Results}
\subsection{Performances of Muffin with GPT-3.5}
\begin{table}[th]
    \centering
    \small
    \renewcommand\arraystretch{1.25}
    \begin{tabular}{c|cccc}
    \hline 
    \text{Model} & \text{emp.} & \text{skill} &    \text{cohr.} & \text{agg.}\\
    \hline
    GPT-3.5 & 92.20 & 84.02 & 91.28 & 75.57   \\
    \hline
    \makecell[c]{Muffin$_{\text{GPT-3.5}}$\\\textit{(w/ one modification)}} & 99.16 & 92.51 &	93.81 & 88.27 \\
    \makecell[c]{Muffin$_{\text{GPT-3.5}}$\\\textit{(w/ two modifications)}} & 99.73 & 96.98 & 95.64 & 93.65  \\
    \hline
    \end{tabular}
    \caption{Performance of Muffin when the base model is GPT-3.5.}
    \label{tab:LLM}
\end{table}
\begin{table}[th]
    \centering
    \small
    \renewcommand\arraystretch{1.25}
    \begin{tabular}{c|ccc}
    \hline 
    Winning Model & \textit{GPT-3.5 }& \textit{Muffin$_{\text{GPT-3.5}}$} & \textit{Tie} \\
    \hline
    Percentages & 8\% & 62\% & 30\% \\
    \hline
    \end{tabular}
    \caption{Human A/B test to compare the helpfulness of GPT-3.5 and Muffin$_{\text{GPT-3.5}}$.}
    \label{tab:LLM_human}
\end{table}
We conducted additional experiments based on an in-context learning GPT-3.5 baseline (the example for each call is randomly selected from the training dataset). 
To implement Muffin, we evaluate the GPT-3.5 generation from multiple facets. 
If the response is identified as unhelpful in any facet, we adjust it from the unhelpful facet(s) until it becomes helpful or we have already attempted modifications twice. 
It is possible that there can be more than two modifications, but we found that the results after two modifications can prove the effectiveness of Muffin. 
In this process, we mitigate the unhelpful responses by pointing out the unhelpful aspects via prompts. 
Unlike the GPT-3.5 baseline, we do not provide a conversation example. 
For evaluation, we used the multifaceted AI feedback module to compute the percentage (\%) of helpful (non-unhelpful) responses in terms of each facet and the aggregated one (similar to the practice in \Cref{tab:facet} left subtable). The results are shown in \Cref{tab:LLM}.

Moreover, we conducted a human evaluation to assess the overall helpfulness of responses. It is important to highlight a key aspect of our experimental design. According to Muffin's design, responses are only modified if they are deemed unhelpful regarding at least one facet. 
Consequently, when 100 instances were randomly chosen for comparison, only 20$\sim$30\% exhibited differing responses between GPT-3.5 and Muffin. 
Thus, we selected 50 instances in which GPT-3.5 and Muffin produced distinct responses. 
We then asked four human annotators to assess each response's overall helpfulness (Helpfulness) within the same context. Results are in \Cref{tab:LLM_human}, where the inter-rater agreement, i.e., Fleiss’ Kappa, is 0.32, indicating a fair agreement.

\subsection{Cases}
For a closer and more detailed examination of our method, we present additional comparisons between the base and Muffin models in \Cref{tab:appn_case}. We include AI feedback for each facet as a reference.

\begin{table*}[htb]
    \small
    \centering
    \renewcommand\arraystretch{1.25}
    \begin{tabular}{m{.11\hsize}<{\centering}m{.41\hsize}|m{.15\hsize}<{\centering}m{.15\hsize}<{\centering}m{.04\hsize}<{\centering}}
    \hline
    \multicolumn{5}{m{.9\hsize}<{\centering}}{Conversation History} \\ 
    \hline
    \multicolumn{5}{m{.95\hsize}}{\textbf{\textit{Supporter}}: Is there anything you'd like to talk to me about?} \\
    \multicolumn{5}{m{.95\hsize}}{\textbf{\textit{Help-seeker}}: I have a friend who knew me since I was little. She is a little older than me. She still treat me like I'm still little. I tried to explain to her that I'm a grown woman and I grew out of lot of things when I was a kid.} \\ 
    \hline
    Model & \makecell[c]{Response} & \textit{\textbf{emp.}} & \textit{\textbf{skill}} & \textit{\textbf{cohr.}}\\ 
    \hline
    Joint & I have had a friend like that before. They are a good friend, and you should feel as though you are a better person. & Weak Empathy & Other & No\\
    Muffin$_{\text{Joint}}$ & I have been in a similar situation. I had a friend who was a little older than me, and I had to deal with the same thing.  & Weak Empathy & Other & Yes \\
    
    \hline
    \multicolumn{5}{m{.9\hsize}<{\centering}}{Conversation History}\\ 
    \hline
    \multicolumn{5}{m{.95\hsize}}{\textbf{\textit{Help-seeker}}: Hello I am very sad today. My partner is cheating on me. I was told by a customer at work. I do not know this person well, yet they knew the person he was cheating on me with. He will not admit it himself.} \\
    \multicolumn{5}{m{.95\hsize}}{\textbf{\textit{Supporter}}: I am sorry if your partner is cheating. So, your partner denies the charge of cheating?} \\
    \multicolumn{5}{m{.95\hsize}}{\textbf{\textit{Help-seeker}}: Yes. He says the person who told me is lying. We are not married but we live together.} \\ 
    \multicolumn{5}{m{.95\hsize}}{\textbf{\textit{Supporter}}: Do you think maybe it was a rumor started by someone who could be jealous? Ash your partner if he loves you enough to propose marriage. If he says no, then ask him why.} \\
    \multicolumn{5}{m{.95\hsize}}{\textbf{\textit{Help-seeker}}: I am not sure if I am okay with giving him an ultimatum. We are still young and although I love being in a relationship, I am unsure about marriage. But I appreciate your suggestions. And you make a great point about the workplace being hostile.} \\ 
    \hline
    Model & \makecell[c]{Response} & \textit{\textbf{emp.}} & \textit{\textbf{skill}} & \textit{\textbf{cohr.}}\\ 
    \hline
    MultiESC & I think you should give him an ultimatum and see if he accepts it. & No Empathy & MI Non-Adherent & Yes \\
    Muffin$_{\text{MultiESC}}$ & Have you spoken to your friends or family about the situation? & Strong Empathy & Other & Yes\\ 

    \hline
    \multicolumn{5}{m{.9\hsize}<{\centering}}{Conversation History} \\ 
    \hline
    \multicolumn{5}{m{.95\hsize}}{\textbf{\textit{Help-seeker}}: I recently broke up with my girlfriend. We had been together for more than 3 years. She said that she simply wasn't in love anymore, and that she didn't have the same feelings she did at the beginning.} \\
    \multicolumn{5}{m{.95\hsize}}{\textbf{\textit{Supporter}}: You still have your whole life ahead of you. Some people don't find a partner until their 40's or even 50's... I think that's what your girl would want you to do as well.} \\
    \multicolumn{5}{m{.95\hsize}}{\textbf{\textit{Help-seeker}}: I disagree with you regarding that maybe that is what she wants me to do. I don't think she cares about me at the slightest anymore. I was a burden for her and will probably keep being it for any woman I meet.} \\ 
    \hline
    Model & \makecell[c]{Response} & \textit{\textbf{emp.}} & \textit{\textbf{skill}} & \textit{\textbf{cohr.}}\\ 
    \hline
    KEMI & That is a great attitude to have. You are a strong person, and I think you will find someone better. & Weak Empathy & MI Non-Adherent & Yes\\
    Muffin$_{\text{KEMI}}$ & I think you are a very good person and I think that you will find someone much better. & Weak Empathy & MI Adherent & Yes\\ 
    \hline
    \end{tabular}
    \caption{Comparison of responses generated by BlenderBot-Joint, MultiESC, and KEMI with their respective Muffin-enhanced counterparts.}
    \label{tab:appn_case}
\end{table*}

\label{sec:appendix}

% This is an appendix.

\end{document}